\documentclass[11pt,a4paper]{article}
\usepackage{times,latexsym}
\usepackage{url}
\usepackage[T1]{fontenc}
\usepackage{graphicx} 
\usepackage{amsmath}      % for math environments like equation
\usepackage{amssymb}      % for \mathbb and other math symbols
\usepackage{bm}           % for bold math (\bm)
\usepackage{multirow}
\usepackage{enumitem}
\usepackage[acceptedWithA]{tacl2021v1}
\usepackage{xspace,mfirstuc,tabulary}

\newif\iftaclinstructions
\taclinstructionsfalse % AUTHORS: do NOT set this to true
\iftaclinstructions
\renewcommand{\confidential}{}
\renewcommand{\anonsubtext}{(No author info supplied here, for consistency with
TACL-submission anonymization requirements)}
\newcommand{\instr}
\fi

\iftaclpubformat % this "if" is set by the choice of options

\else

\fi

\title{EVENT5Ws: A Large Dataset for Open-Domain Event Extraction from Documents}

\author{
  Praval Sharma$^{1}$ \hspace{1em}
  Ashok Samal$^{2}$ \hspace{1em}
  Leen-Kiat Soh$^{2}$ \hspace{1em}
  Deepti Joshi$^{3}$ \\
  $^{1}$University of Nebraska Omaha, USA \\
  $^{2}$University of Nebraska--Lincoln, USA \\
  $^{3}$The Citadel, USA
}

\date{}

\begin{document}
\maketitle
\begin{abstract}
   Event extraction identifies the central aspects of events from text. It supports event understanding and analysis, which is crucial for tasks such as informed decision-making in emergencies. Therefore, it is necessary to develop automated event extraction approaches. However, existing datasets for algorithm development have limitations, including limited coverage of event types in closed-domain settings and a lack of large, manually verified dataset in open-domain settings. To address these limitations, we create EVENT5Ws, a large, manually annotated, and statistically verified open-domain event extraction dataset. We design a systematic annotation pipeline to create the dataset and provide empirical insights into annotation complexity. Using EVENT5Ws, we evaluate state-of-the-art pre-trained large language models and establish a benchmark for future research. We further show that models trained on EVENT5Ws generalize effectively to datasets from different geographical contexts, which demonstrates its potential for developing generalizable algorithms. Finally, we summarize the lessons learned during the dataset development and provide recommendations to support future large-scale dataset development.
\end{abstract}

%\iftaclpubformat
\section{Introduction}
An event is an activity that occurs at a particular location and time \cite{allanTopicDetectionTracking1998}. It is represented by different aspects that provide insights into its location, time, participants, and the underlying cause \cite{yuSpatiotemporalEventDetection2020}. Event extraction aims to identify these central aspects of events from text. It is an essential step toward understanding and analyzing events, and is necessary for various tasks, such as improving situational awareness, facilitating effective management and mitigation activities in emergencies, and helping stakeholders in decision-making. Therefore, it is necessary to develop algorithms that can automatically extract events.

\setlength{\parskip}{0pt}A standardized and well-vetted dataset is central to the systematic development and accurate evaluation of algorithms. However, existing event extraction datasets have several limitations. Sentence-level closed-domain event-extraction datasets \cite{getmanOverviewLinguisticResources2017, parekhGENEVABenchmarkingGeneralizability2023} are unsuitable because events are generally described across multiple sentences within a document \cite{tongDocEELargeScaleFinegrained2022} and using them may result in extraction of incomplete information. Document-level closed-domain event extraction datasets \cite{liDocumentLevelEventArgument2021, tongDocEELargeScaleFinegrained2022} address this issue. However, they are created based on predefined event schemas designed for specific event types (e.g., 59 event types such as ‘Conflict-Attack,’ ‘Justice-Investigate,’ and ‘Disaster-AccidentCrash’ in \citet{tongDocEELargeScaleFinegrained2022}). Since real-world events are inherently diverse and evolving, enumerating all possible event types is impractical. As a result, algorithms trained on such datasets struggle to generalize to unseen event types. This motivates the need for open-domain event extraction datasets, which support development of algorithms that can extract unconstrained event types and are essential for natural language understanding \cite{arakiOpenDomainEventDetection2018}. However, existing open-domain datasets \cite{liuOpenDomainEvent2019, veysehAugmentingOpenDomainEvent2021} contain a relatively small number of documents with manually verified annotations (e.g., 680 in  \citet{liuOpenDomainEvent2019}), which may not be sufficient for developing advanced deep learning-based approaches. While synthetic data generation using generative models offers a potential approach for increasing dataset size, prior research shows that such models remain unreliable for event extraction as they struggle to preserve coherence over long document context \cite{cao5W1HExtractionLarge2024}. Therefore, it is necessary to create large and manually verified open-domain event extraction datasets.

In this research, we present a large dataset, EVENT5Ws, for document-level open-domain event extraction using manual annotation and validated through an inter-coder reliability (ICR) measure. Documents, particularly news reports, are one of the primary sources of information on events \cite{ritterOpenDomainEvent2012}. We therefore leverage them to create EVENT5Ws. We employ the 5Ws framework and annotate the five key aspects of events, i.e., where, when, what, who, and why, in documents to create the dataset. We use the 5Ws framework because it forms the foundation for reporting events in news reports \cite{harrowerReportingPracticalGuide2010}, is applicable to unconstrained event types, and has been used by several prior open-domain event extraction studies \cite{hamborgGiveme5W1HUniversalSystem2019, liuOpenDomainEvent2019}. We adopt the one-event-per-document approach of \citet{tongDocEELargeScaleFinegrained2022} and annotate the 5Ws for the main events described in documents. Additionally, we provide a discussion on the lessons learned during the annotation process that may benefit future similar large-scale dataset development efforts and show that contextual familiarity of coders plays a critical role in annotation efficiency. We also evaluate several state-of-the-art (SOTA) large language models (LLMs) on EVENT5Ws to establish baselines and provide a comprehensive benchmark. The evaluation highlights the challenging nature of open-domain event extraction as the models struggle to perform effectively and the need for development of specialized algorithms, which requires large datasets such as EVENT5Ws. Through additional experiments, we demonstrate that models trained on EVENT5Ws generalize effectively across datasets from diverse geographical and textual contexts, which highlights the dataset’s potential for developing robust open-domain event extraction algorithms. 

\setlength{\parskip}{0pt}Our contributions are: (1) One of the largest open-domain event extraction datasets, EVENT5Ws, developed using manual annotation and verified through ICR, (2) a systematic annotation pipeline for large-scale event extraction dataset development, with empirical insights into annotation complexity, and (3) evaluation of SOTA pre-trained LLMs and establishment of a benchmark for future open-domain event-extraction research.

Note that, in general, documents describe newsworthy, recent events central to their content, i.e., the main events, along with background events that provide supporting information \cite{grimesThreadDiscourse1975}. Though background events provide contextual information, documents typically center around the main events \cite{dijkNewsDiscourse1988}. Thus, extracting the main events from documents is essential for news discourse and comprehension. Several studies on news discourse analysis \cite{choubeyDiscourseFunctionEvent2020, upadhyayMakingNewsIdentifying2016} and event-centric news clustering \cite{zhangEnhancingEventcentricNews2025} have used main events as the core unit of analysis. Additionally, in real-world emergency scenarios such as natural disasters or civil unrest, information about the main event is critical for effective response planning and monitoring. This motivates our one-event-per-document assumption, under which we annotate the 5Ws for the main event. We consider the most central event agreed by our coders on a document as the main event.
 
\section{Related Work}

\textbf{Closed-Domain Event Extraction Dataset: } These datasets are developed based on predefined event schemas designed for specific event types. For example, an event of type ‘Conflict-Attack’ will have an event schema consisting of arguments such as ‘attacker,’ ‘target,’ and ‘instrument.’ They can be divided into sentence-level event extraction \cite{getmanOverviewLinguisticResources2017, songLightRichERE2015} and document-level event extraction \cite{ebnerMultiSentenceArgumentLinking2020, tongDocEELargeScaleFinegrained2022, xianDLEEDatasetChinese2024} based on their usage of either a sentence or a document. Information about an event in a document is generally mentioned across multiple sentences \cite{ tongDocEELargeScaleFinegrained2022}. Thus, using sentence-level datasets can result in the extraction of incomplete information about events. The document-level datasets provide opportunities to extract complete information about events. However, since existing document-level datasets are developed using predefined event schemas tailored to specific event types (e.g., 139 types of events in \citet{ebnerMultiSentenceArgumentLinking2020} and 59 in \citet{tongDocEELargeScaleFinegrained2022}), algorithms trained on such datasets rarely generalize to events not covered in the predefined schemas. Unlike these datasets, EVENT5Ws is based on the 5Ws framework that generalizes to all event types and supports the development of approaches for extracting unconstrained event types.

\textbf{Open-Domain Event Extraction Dataset: } These datasets \cite{arakiOpenDomainEventDetection2018, cao5W1HExtractionLarge2024, hamborgGiveme5W1HUniversalSystem2019, minardMEANTIMENewsReaderMultilingual2016, simsLiteraryEventDetection2019} are created without predefined event schemas. As a result, they facilitate the development of approaches capable of extracting unconstrained event types, which is essential for advancing natural language understanding and downstream applications \cite{arakiOpenDomainEventDetection2018}. However, the number of manually annotated documents in these datasets is limited (e.g., 100 in \citet{arakiOpenDomainEventDetection2018} and 680 in \citet{liuOpenDomainEvent2019}). This relatively small scale may limit their utility for developing robust deep learning approaches to open-domain event extraction, as it can restrict model generalization and negatively impact performance.

To increase the dataset size, several prior works have explored data augmentation using methods such as knowledge bases \cite{arakiOpenDomainEventDetection2018} and synthetic data generation using generative models \cite{veysehAugmentingOpenDomainEvent2021}. However, these unsupervised and manually unverified augmentations raise concerns about the quality of the generated data \cite{tongDocEELargeScaleFinegrained2022}. Additionally, even when synthetic data is generated using more recent large-scale generative models, research indicates that such models still fall short of the reliability required for high-quality dataset development, particularly for event extraction, as they struggle to produce coherent outputs over long document contexts \cite{cao5W1HExtractionLarge2024}. In this work, we create a large-scale, manually annotated, and statistically validated dataset, EVENT5Ws, that follows the open-domain event extraction paradigm.

\section{Development of EVENT5Ws}
\subsection{Coder Selection and Recruitment Process}
The coders used for an annotation task significantly impact annotation time, quality, and cost \cite{hovyExperimentsCrowdsourcedReannotation2014}. Therefore, coder selection is an important aspect for any annotation task. Experts can produce high quality annotations but are expensive and not scalable \cite{sorokinUtilityDataAnnotation2008}. Crowdsourcing is a cheap and scalable alternative \cite{snowCheapFastIt2008}. However, it does not ensure high-quality annotations \cite{danielQualityControlCrowdsourcing2018} and may be negatively impacted by monetary incentives \cite{drutsaCrowdsourcingPracticeEfficient2020}. Citizen science is another approach that uses coders who do not work for monetary gain but are intrinsically motivated. Due to this intrinsic motivation, these coders may invest more time and effort to produce accurate annotations \cite{leeAnnotationCurriculaImplicitly2022}. In general, intrinsically motivated coders are recruited through various channels such as social media, mailing lists, and university courses \cite{klieLessonsLearnedCitizen2023, tsuengCitizenScienceMining2016}. Among these channels, university students are high performing coders capable of producing quality annotations \cite{klieLessonsLearnedCitizen2023}. Based on this finding, we also employ students to create EVENT5Ws.

To recruit them, we advertise the opportunity across various university courses, departmental mailing lists, and student associations to reach a broad pool of candidates. We communicate the complete details of the study, including the tasks and compensation, to all potential coders. Additionally, we make the participation of students voluntary to meet the ethical standards, compensate them at the basic minimum pay rate set by the university for student workers, and take measures to preserve privacy and confidentiality by not disclosing their personal details publicly. 

\subsection{Annotation Platform}
Performing annotation with multiple coders requires a platform that supports annotation, monitoring, and data handling capabilities. However, several existing annotation platforms (e.g., Amazon Mechanical Turk\footnote{https://www.mturk.com/}, Prodigy\footnote{https://prodi.gy/}) are not freely available. Subscribing to and using them increases the already high cost of annotation. Therefore, we use Dataturks\footnote{https://github.com/DataTurks}, a freely available open-source web-based application that allows various functionalities for easy and interactive text annotation (see Appendix \ref{appen1}), to create EVENT5Ws. We download a version and host locally on a server. This in-house setup enables effective monitoring and handling of the coders and their annotations.

\subsection{Annotation Guidelines}
Clear and unambiguous guidelines are essential for effective manual annotation to create a large dataset. In addition, providing illustrative examples helps coders better understand the task and produce accurate annotations. In this study, we create clear guidelines along with specific examples for event annotation. To do this, a set of four experts (researchers with extensive experience in event extraction) first analyzed a set of randomly selected documents and identified guidelines for locating and annotating the events. After that, feedback taken from annotators during pilot testing was used to refine the guidelines. The final guidelines are described in Appendix \ref{appen2}.

\subsection{Resolution Policy}
The resolution policy serves as the guidelines to systematically handle disagreements among coders and ensure that the final dataset is accurate and reliable. Our resolution policy is as follows.
\begin{itemize}[nosep] 
    \item When all coders agree (i.e., unanimous agreement), the annotation is considered the gold standard.
    \item When only two coders agree (i.e., partial agreement), the lone coder with the disagreement is asked to reconsider and annotate again. In this process, the lone coder will be provided with the annotations from the other coders to provide contextual information about the previous annotations so that it would benefit the coder in making a decision. Subsequently, if all the coders agree, then the annotation is considered the gold standard. If the differences still exist, an expert determines the gold standard. Note that the expert can either go along with one of the annotations from the coders or select a different word or phrase as the gold standard.
    \item When all coders disagree (i.e., complete disagreement), all of them redo the annotation, and the process is repeated until at least two coders agree on the annotations.
\end{itemize}

\subsection{Dataset Construction Process}
\begin{figure}[t]
\centering
\includegraphics[width=\linewidth]{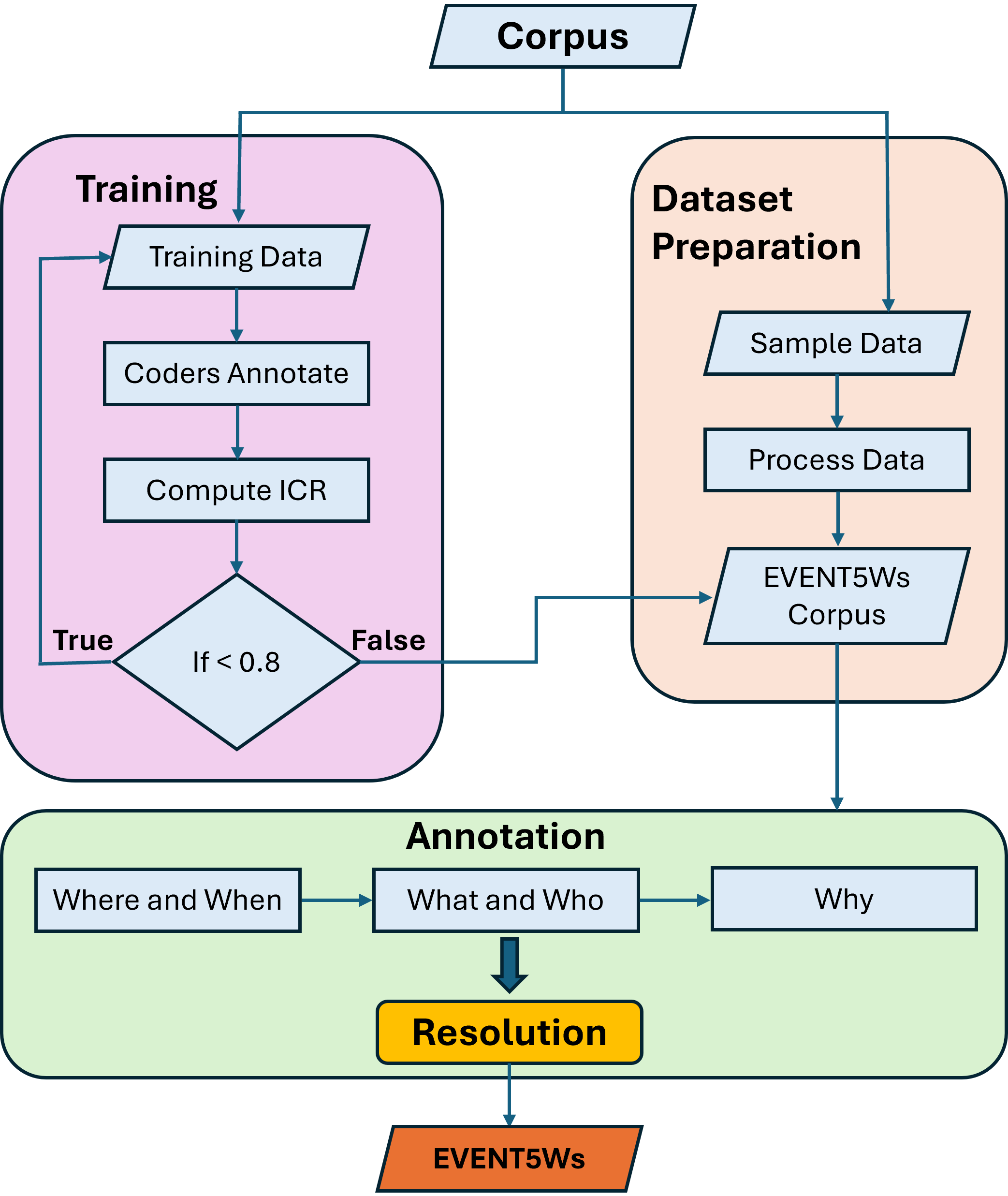}
\caption{The overall schematic of our approach to developing EVENT5Ws.}
\label{figure1}
\end{figure}

Our approach to creating EVENT5Ws consists of four different steps: (1) Training, (2) Dataset preparation, (3) Annotation, and (4) Resolution of the disagreements in the annotations (see Figure \ref{figure1}). These four steps are described below.

\textbf{Training: }This step involves training the coders to understand the annotation guidelines, become familiar with the annotation platform, and consistently produce quality annotations. It begins by providing the annotation guidelines and a demo of the annotation process in the annotation platform. Next, a small set of documents are randomly sampled from the corpus and provided to the coders for annotation. The ICR is then computed based on these annotations and the disagreements are discussed with the coders. This approach clarifies ambiguous cases and helps improve the coders’ understanding of the annotation guidelines. The training is repeated until an acceptable ICR is achieved. 

\textbf{Dataset Preparation: }This step involves selecting a set of documents to create the EVENT5Ws corpus (see Section 4.2). The documents are selected using random sampling from newspapers with diverse scopes (e.g., general news, financial news) and spatial coverage (e.g., national, regional). This helps enrich the diversity of the dataset in terms of writing styles and content. The selected documents are preprocessed to remove noise, such as html tags, extra white space and line breaks, and encoding artifacts, before being included in the EVENT5Ws corpus.

\textbf{Annotation: }In this step, coders annotate the 5Ws for the main events in the EVENT5Ws corpus. Where and When are generally represented by location and time markers \cite{cao5W1HExtractionLarge2024}. Thus, annotating them is relatively easier. In contrast, annotating What and Who is more challenging as What is typically represented by generic words or phrases that require global contextual and semantic understanding \cite{hamborgGiveme5W1HUniversalSystem2019}, while Who may involve multiple actors. Why is the most challenging as it requires deeper contextual and sematic understanding of documents and is frequently implicit or absent \cite{hamborgGiveme5W1HUniversalSystem2019}. 

To account for these varying levels of difficulty, the annotation task is divided into three phases, i.e., Where and When in the first phase, Who and What in the second, and Why in the third. This phased approach is based on the principles outlined in \citet{leeAnnotationCurriculaImplicitly2022}, which argue that introducing annotation tasks in increasing order of difficulty improves the accuracy and consistency while reducing annotation time. Additionally, to facilitate better time and resource management, enable regular monitoring and discussion, and help coders maintain focus, which enhances annotation quality, the reports are organized into batches of 1,000 reports in each phase.

\textbf{Resolution: }This step involves resolving the differences in the annotations from the coders governed by the resolution policy. At the end of this step, we produce a large and reliable dataset, EVENT5Ws, for open-domain event extraction.

\section{Dataset and Annotation}
\subsection{Corpus}
We collected news reports from seven different English-language newspapers from India (Times of India, The Hindu, The Pioneer, Economic Times, Assam Tribune, Kashmir Observer, and Incredible Orissa) to create a large corpus. Of the seven newspapers, four are national and three are regional. They have diverse foci and journalistic writing styles. For example, the Economic Times primarily focuses on financial news whereas The Hindu covers a broad range of generic news (e.g., current affairs, technology, health, etc.). Additionally, the regional newspapers cover different geographic regions within India. For example, Assam Tribune serves the Northeast region, whereas Incredible Orissa serves the Eastern region. Overall, these diverse sources contribute to the diversity of the corpus and are selected based on their coverage and the availability of their archives. In total, the corpus consists of 5.24 million news reports (see Appendix \ref{appen3} for corpus statistics across sources).

\subsection{EVENT5Ws Corpus}
The EVENT5Ws corpus is created by randomly sampling 10,000 reports from the corpus described in Section 4.1. This sampling covers the years 2015 to 2019 to match the intersection of the temporal scopes of all newspapers. The number of reports sampled from each newspaper is proportional to its contribution to the overall corpus to ensure balanced representation (see Table \ref{table5}).

According to the inverse pyramid concept, important information is usually mentioned at the beginning of reports \cite{harrowerReportingPracticalGuide2010}. Since main events are central to their content \cite{dijkNewsDiscourse1988}, they are likely to appear early. Manual annotation is expensive and time-consuming, and because we wanted to include many reports for diversity while balancing cost, we chose to truncate them to include parts where main events occur frequently. We undertook a preliminary experiment on 74 manually annotated reports and found that main events appeared in the first five sentences in 97.3\% of cases. This observation aligns with \citet{ebnerMultiSentenceArgumentLinking2020}, who found that most of an event’s arguments in reports appeared within a five-sentence window. Based on this, we included the title and first five sentences of reports in the EVENT5Ws corpus.

\subsection{Coder Recruitment and Training Summary}
After advertising the opportunity, we received interest from 6 students. In general, greater the number of coders, higher the credibility of the ICR \cite{krippendorffAgreementInformationReliability2011}. However, recruiting many coders is challenging and costly. Therefore, we decided to recruit three coders to bolster the credibility of annotations while minimizing cost. We selected the three (one undergraduate freshman and two second-year graduate students) based on their familiarity with the geographic and cultural context of our dataset (i.e., India) so that they could accurately and consistently identify the main events.

Three rounds of training were conducted before obtaining an acceptable ICR ($\ge 0.80$) for all the 5Ws, for all 5Ws, measured using Krippendorff’s alpha \cite{krippendorffReliabilityMultiValuedCoding2016}. In each round, coders annotated seven reports, one from each of the seven newspapers, not included in the EVENT5Ws. Annotation consistency improved for all 5Ws from the first to the third round corpus (see Appendix \ref{appen4} for the ICR values computed across the training rounds). In the first round, coders were unfamiliar with the task, resulting in lower ICR values. By the third round, they became consistent. Prior knowledge of the geographic and cultural context of the dataset also helped the coders and contributed to the high ICR reached after the third round.

Computing ICR using Krippendorff’s alpha relies on exact matches (EMs) between annotations. The annotations for Where, When, and Who typically consist of location, person, organization, or a specific date and temporal adverbs. As a result, it is straightforward to compute ICR using EM. However, annotations for What and Why can include an individual word or a phrase, where two annotations that differ lexically may be semantically similar. For example, ‘protest’ and ‘held a protest march’ are What annotations in a report during training. Though they conceptually express the same phenomenon and denote the coders’ agreement, ICR based on EM gives a score of zero. Thus, computing ICR using EM, particularly for What and Why, is not practical.

Computing ICR using Krippendorff’s alpha relies on exact matches (EMs) between annotations. The annotations for Where, When, and Who typically consist of location, person, organization, or a specific date and temporal adverbs. As a result, it is straightforward to compute ICR using EM. However, annotations for What and Why can include an individual word or a phrase, where two annotations that differ lexically may be semantically similar. For example, ‘protest’ and ‘held a protest march’ are What annotations in a report during training. Though they conceptually express the same phenomenon and denote the coders’ agreement, ICR based on EM gives a score of zero. Thus, computing ICR using EM, particularly for What and Why, is not practical. 

To better measure agreement (or disagreement) among the coders for What and Why, we used spaCy\footnote{https://spacy.io/}, a widely used NLP tool, to compute the semantic similarity between the annotations and identify whether they can be considered a match for computing ICR. We undertook a preliminary experiment to determine a suitable similarity score threshold for assessing the annotations. A manual analysis of the similarity scores computed between What and Why annotations from our coders in a set of 74 reports demonstrated that conceptually similar annotations had a similarity score of 0.61 or higher. Therefore, we considered annotations with similarity score of 0.61 or higher as a match and computed the ICR in different training rounds for What and Why.

Annotating a large dataset requires significant time and it is challenging to retain coders throughout the annotation process. We had to replace two coders as they graduated after completing Where and When annotations in the first phase. We recruited and trained two new coders (both second-year graduate students) using the same approach as the initial coders.

\subsection{Annotation and Resolution Summary}
Table \ref{table1} presents the number of reports where coders agreed on their annotations during the annotation step. Intuitively, it is relatively easier to identify Where, When, and Who as they are represented by place names, temporal expressions, and names of persons or organizations. As a result, they have higher agreements. In contrast, identifying What and Why is difficult as they lack specific textual markers and require deeper contextual understanding. In addition, event’s causal aspect is often missing or implicit, which increases Why’s ambiguity. Due to this, What and Why have relatively higher disagreements.

The EVENT5Ws corpus was divided into batches of 1,000 reports per phase of the annotation step (see Section 3.5) to facilitate regular monitoring and discussion. To track coder performance over time, we computed the number of disagreements within each batch (see Tables \ref{table2} and \ref{table3}). The first batch had the most disagreements across all 5Ws, likely due to coder inexperience. In the second batch, disagreements decreased noticeably, which indicates quick adaptation of coders. However, in subsequent batches, number of disagreements remained consistently similar. Factors such as source newspapers, writing style, and fatigue in coders may have contributed to this stagnation. 

To understand the annotation complexity, we tracked the average time taken to annotate each W (see Table \ref{table4}). Annotating Why consumed the most time, which reflects the difficulty of identifying the causal aspect of events. In contrast, Where and When took the least time as location and time specific markers are easier to identify. Coders’ familiarity with the geographic and cultural context may also have helped in recognizing these markers efficiently.

After completing the annotation step, we conducted a resolution step to address disagreements, where coders re-annotated the reports they disagreed on per the resolution policy (see Section 3.4). Table \ref{table4} shows the average time per report during this step. Coders were provided with all prior annotations to ensure they had proper context for their revisions. This likely facilitated a more informed resolution and improved efficiency as the average time per report decreased compared to the annotation step for all Ws.

Any unresolved disagreements were further resolved by two experts, i.e., researchers with knowledge in event extraction, per the resolution policy. They resolved 655 reports. At the end, we created the EVENT5Ws dataset containing 10,000 reports for open-domain event extraction (see Appendix \ref{appen1} for dataset details).

\begin{table}[t]
\centering
\resizebox{\columnwidth}{!}{%
\begin{tabular}{l|c|c|c}
\hline
\multicolumn{1}{c|}{\textbf{5Ws}} & \textbf{All Agree} & \textbf{Only   Two Agree} & \textbf{All Disagree} \\ \hline
\textbf{Where} & 6880 (68.8\%) & 2943 (29.4\%) & 177 (1.8\%) \\ 
\textbf{When} & 7571 (75.7\%) & 1865 (18.7\%) & 52 (0.5\%) \\ 
\textbf{Who} & 8712 (87.1\%) & 927 (9.3\%) & 361 (3.6\%) \\ 
\textbf{What} & 5166 (51.7\%) & 2926 (29.3\%) & 1908 (19.1\%) \\ 
\textbf{Why} & 4417 (44.2\%) & 3516 (35.2\%) & 2067 (20.7\%) \\ \hline
\end{tabular}}
\caption{Summary of agreements and disagreements among coders on their annotations.}
\label{table1}
\end{table}

% \usepackage{multirow}
\begin{table}[t]
\centering
\resizebox{\columnwidth}{!}{%
\begin{tabular}{l|cccccccccc}
\hline
\multicolumn{1}{c|}{\multirow{2}{*}{\textbf{5Ws}}} & \multicolumn{10}{c}{\textbf{Batch}} \\ \cline{2-11} 
\multicolumn{1}{c|}{} & \multicolumn{1}{c|}{\textbf{1}} & \multicolumn{1}{c}{\textbf{2}} & \multicolumn{1}{c|}{\textbf{3}} & \multicolumn{1}{c|}{\textbf{4}} & \multicolumn{1}{c}{\textbf{5}} & \multicolumn{1}{c|}{\textbf{6}} & \multicolumn{1}{c|}{\textbf{7}} & \multicolumn{1}{c|}{\textbf{8}} & \multicolumn{1}{c|}{\textbf{9}} & \textbf{10} \\ \hline
\textbf{Where} & \multicolumn{1}{c|}{351} & \multicolumn{1}{c|}{315} & \multicolumn{1}{c|}{295} & \multicolumn{1}{c|}{285} & \multicolumn{1}{c|}{273} & \multicolumn{1}{c|}{279} & \multicolumn{1}{c|}{274} & \multicolumn{1}{c|}{280} & \multicolumn{1}{c|}{288} & 294 \\ 
\textbf{When} & \multicolumn{1}{c|}{240} & \multicolumn{1}{c|}{197} & \multicolumn{1}{c|}{188} & \multicolumn{1}{c|}{176} & \multicolumn{1}{c|}{159} & \multicolumn{1}{c|}{165} & \multicolumn{1}{c|}{183} & \multicolumn{1}{c|}{177} & \multicolumn{1}{c|}{189} & 191 \\ 
\textbf{Who} & \multicolumn{1}{c|}{145} & \multicolumn{1}{c|}{99} & \multicolumn{1}{c|}{86} & \multicolumn{1}{c|}{83} & \multicolumn{1}{c|}{82} & \multicolumn{1}{c|}{80} & \multicolumn{1}{c|}{85} & \multicolumn{1}{c|}{86} & \multicolumn{1}{c|}{88} & 93 \\ 
\textbf{What} & \multicolumn{1}{c|}{452} & \multicolumn{1}{c|}{285} & \multicolumn{1}{c|}{270} & \multicolumn{1}{c|}{269} & \multicolumn{1}{c|}{263} & \multicolumn{1}{c|}{265} & \multicolumn{1}{c|}{269} & \multicolumn{1}{c|}{280} & \multicolumn{1}{c|}{284} & 289 \\ 
\textbf{Why} & \multicolumn{1}{c|}{549} & \multicolumn{1}{c|}{385} & \multicolumn{1}{c|}{350} & \multicolumn{1}{c|}{330} & \multicolumn{1}{c|}{312} & \multicolumn{1}{c|}{303} & \multicolumn{1}{c|}{305} & \multicolumn{1}{c|}{322} & \multicolumn{1}{c|}{325} & 335 \\ \hline
\end{tabular}}
\caption{Number of reports with only one coder in disagreement.}
\label{table2}
\end{table}

% \usepackage{multirow}
\begin{table}[t]
\centering
\resizebox{\columnwidth}{!}{%
\begin{tabular}{l|cccccccccc}
\hline
\multicolumn{1}{c|}{\multirow{2}{*}{\textbf{5Ws}}} & \multicolumn{10}{c}{\textbf{Batch}} \\ \cline{2-11} 
\multicolumn{1}{c|}{} & \multicolumn{1}{c|}{\textbf{1}} & \multicolumn{1}{c|}{\textbf{2}} & \multicolumn{1}{c|}{\textbf{3}} & \multicolumn{1}{c|}{\textbf{4}} & \multicolumn{1}{c|}{\textbf{5}} & \multicolumn{1}{c|}{\textbf{6}} & \multicolumn{1}{c|}{\textbf{7}} & \multicolumn{1}{c|}{\textbf{8}} & \multicolumn{1}{c}{\textbf{9}} & \textbf{10} \\ \hline
\textbf{Where} & \multicolumn{1}{c|}{29} & \multicolumn{1}{c|}{22} & \multicolumn{1}{c|}{19} & \multicolumn{1}{c|}{17} & \multicolumn{1}{c|}{15} & \multicolumn{1}{c|}{14} & \multicolumn{1}{c|}{13} & \multicolumn{1}{c|}{15} & \multicolumn{1}{c|}{16} & 17 \\ 
\textbf{When} & \multicolumn{1}{c|}{11} & \multicolumn{1}{c|}{8} & \multicolumn{1}{c|}{5} & \multicolumn{1}{c|}{6} & \multicolumn{1}{c|}{4} & \multicolumn{1}{c|}{3} & \multicolumn{1}{c|}{3} & \multicolumn{1}{c|}{4} & \multicolumn{1}{c|}{6} & 2 \\ 
\textbf{Who} & \multicolumn{1}{c|}{53} & \multicolumn{1}{c|}{35} & \multicolumn{1}{c|}{34} & \multicolumn{1}{c|}{32} & \multicolumn{1}{c|}{30} & \multicolumn{1}{c|}{31} & \multicolumn{1}{c|}{33} & \multicolumn{1}{c|}{34} & \multicolumn{1}{c|}{40} & 39 \\ 
\textbf{What} & \multicolumn{1}{c|}{280} & \multicolumn{1}{c|}{180} & \multicolumn{1}{c|}{175} & \multicolumn{1}{c|}{174} & \multicolumn{1}{c|}{170} & \multicolumn{1}{c|}{172} & \multicolumn{1}{c|}{182} & \multicolumn{1}{c|}{192} & \multicolumn{1}{c|}{190} & 193 \\ 
\textbf{Why} & \multicolumn{1}{c|}{320} & \multicolumn{1}{c|}{250} & \multicolumn{1}{c|}{220} & \multicolumn{1}{c|}{200} & \multicolumn{1}{c|}{185} & \multicolumn{1}{c|}{175} & \multicolumn{1}{c|}{170} & \multicolumn{1}{c|}{180} & \multicolumn{1}{c|}{182} & 185 \\ \hline
\end{tabular}}
\caption{Number of reports with disagreement among all three coders.}
\label{table3}
\end{table}

% \usepackage{multirow}
\begin{table}[t]
\centering
\resizebox{\columnwidth}{!}{%
\begin{tabular}{l|cc}
\hline
\multicolumn{1}{c|}{\multirow{2}{*}{\textbf{5Ws}}} & \multicolumn{2}{c}{\textbf{Average   Time (Seconds)}} \\ \cline{2-3} 
\multicolumn{1}{c|}{} & \multicolumn{1}{c|}{\textbf{Annotation   Step}} & \textbf{Resolution   Step} \\ \hline
\textbf{Where and When} & \multicolumn{1}{c|}{55.25} & 44.36 \\ 
\textbf{Who and What} & \multicolumn{1}{c|}{68.55} & 56.74 \\ 
\textbf{Why} & \multicolumn{1}{c|}{62.32} & 53.27 \\ \hline
\end{tabular}}
\caption{Average time taken per report by coders in the annotation and resolution steps.}
\label{table4}
\end{table}
\section{Experiments}
\subsection{Dataset and Evaluation}
We randomly split EVENT5Ws into training, validation, and test sets containing 8,000, 1,000, and 1,000 documents, respectively, for our experiments. Table \ref{table5} shows the distribution of documents across these splits. Each split contains reports from all seven newspapers, with approximately proportional representation to ensure balanced coverage and support robust training and reliable evaluation. The training and validation sets are for model training and the test set is for evaluation.

Following prior work on event extraction \cite{duEventExtractionAnswering2020, tongDocEELargeScaleFinegrained2022}, we report the results using precision (P), recall (R), and F1 score computed separately for each of the 5W classes using EM, where a prediction is considered correct if it exactly matches the corresponding gold annotation. Additionally, to account for the Ws such as What and Why, where semantically equivalent outputs may differ in surface form, we also report results using an overlap-based metric ROUGE-L \cite{linROUGEPackageAutomatic2004}. ROUGE-L precision, recall, and F1 score are computed per document and averaged over all documents in the test set.

% \usepackage{multirow}
\begin{table}[t]
\centering
\resizebox{\columnwidth}{!}{%
\begin{tabular}{l|cccc}
\hline
\multicolumn{1}{c|}{\multirow{2}{*}{\textbf{Newspaper}}} & \multicolumn{4}{c}{\textbf{\# Reports}} \\ \cline{2-5} 
\multicolumn{1}{c|}{} & \multicolumn{1}{c|}{\textbf{Train}} & \multicolumn{1}{c|}{\textbf{Validation}} & \multicolumn{1}{c}{\textbf{Test}} & \textbf{TOTAL} \\ \hline
\textbf{Times   of India} & \multicolumn{1}{c|}{3573} & \multicolumn{1}{c|}{445} & \multicolumn{1}{c|}{460} & 4478 \\ 
\textbf{The   Hindu} & \multicolumn{1}{c|}{1893} & \multicolumn{1}{c|}{228} & \multicolumn{1}{c|}{246} & 2367 \\ 
\textbf{The   Pioneer} & \multicolumn{1}{c|}{503} & \multicolumn{1}{c|}{62} & \multicolumn{1}{c|}{61} & 626 \\ 
\textbf{Economic   Times} & \multicolumn{1}{c|}{1821} & \multicolumn{1}{c|}{239} & \multicolumn{1}{c|}{204} & 2264 \\ 
\textbf{Assam   Tribune} & \multicolumn{1}{c|}{170} & \multicolumn{1}{c|}{21} & \multicolumn{1}{c|}{26} & 217 \\ 
\textbf{Kashmir   Observer} & \multicolumn{1}{c|}{36} & \multicolumn{1}{c|}{4} & \multicolumn{1}{c|}{2} & 42 \\ 
\textbf{Incredible   Orissa} & \multicolumn{1}{c|}{4} & \multicolumn{1}{c|}{1} & \multicolumn{1}{c|}{1} & 6 \\ \hline
\textbf{TOTAL} & \multicolumn{1}{c|}{8000} & \multicolumn{1}{c|}{1000} & \multicolumn{1}{c|}{1000} & 10000 \\ \hline
\end{tabular}}
\caption{Number of reports from each newspaper in the training, validation, and test set in EVENT5Ws.}
\label{table5}
\end{table}

\subsection{Baselines}
We evaluate several SOTA pretrained LLMs, including Gemma 3 (12B Instruct), Llama 3.1 (8B and 70B Instruct\footnote{We use 8-bit quantization due to memory constraints.}), Qwen 3 (4B Instruct and 32B Thinking), Mistral v0.3 (7B Instruct), and T5 Large, on the task of main event 5Ws extraction under zero-shot and five-shot prompting (see Appendix \ref{appen5} for example prompts). These models represent a diverse set of architectures and capabilities, including instruction-tuned and reasoning, which enables a comprehensive evaluation. As LLM-based open-domain event extraction methods are not publicly available and remain limited due to the lack of large-scale annotated datasets, direct comparison is not feasible. Therefore, we use these widely adopted LLMs as strong baselines. During inference, for all models, we use the maximum new tokens parameter of 512 with temperature set to 0.1 and sampling enabled to ensure fair evaluation.

\subsection{Result}
Table \ref{table6} presents the performance of all models on the test set under zero-shot and five-shot prompting using EM. Performance varies significantly across the 5Ws and no single model consistently outperforms others. All models achieve relatively higher scores on Where, When, and Who, while performance on What and Why is consistently lower. Where, When, and Who are typically represented using named entities and thus are relatively easier to identify as reflected in Table \ref{table6}. This observation is consistent with our annotation process, where these Ws required less annotation time and exhibited higher coder agreements (see Section 4.4). In contrast, What and Why are generally represented using more diverse and unstructured language without explicit textual markers. As a result, extracting them requires deeper contextual understanding and reasoning across the document. The lower performance on these Ws highlights the increased difficulty of extracting them.

Five-shot prompting improved performance over zero-shot across all models and Ws. This indicates that in-context examples provide useful guidance. However, the overall gains are modest. Additionally, larger models (e.g., Llama 70B) or models with reasoning capability (e.g., Qwen 3 32B) do not lead to better performance. This shows the limitations of current LLMs in handling document-level event extraction that requires understanding of the global context and capturing the structural semantics of event-related tokens.

Table \ref{table7} shows that evaluation using ROUGE-L produces higher scores than EM across all models and Ws. This indicates that models often generate partially correct or semantically similar outputs that differ in surface form from the gold annotations. However, substantial room for improvement remains as the models struggle to produce precise and consistent outputs. 

Overall, despite their advanced capabilities, the evaluated LLMs struggle to extract accurate main event 5Ws for most documents in EVENT5Ws. This demonstrates the complexity of event extraction and the need for more specialized approaches beyond off-the-shelf models. Resources such as EVENT5Ws can support the development of such approaches, and the results in Tables \ref{table6} and \ref{table7} provide a meaningful benchmark.

% \usepackage{multirow}
\begin{table*}[t]
\centering
\resizebox{\textwidth}{!}{
\begin{tabular}{l|ccc|ccc|ccc|ccc|ccc}
\hline
\multicolumn{1}{c|}{\multirow{2}{*}{\textbf{Model}}} & \multicolumn{3}{c|}{\textbf{Where}} & \multicolumn{3}{c|}{\textbf{When}} & \multicolumn{3}{c|}{\textbf{What}} & \multicolumn{3}{c|}{\textbf{Who}} & \multicolumn{3}{c}{\textbf{Why}} \\ \cline{2-16} 
\multicolumn{1}{c|}{} & \multicolumn{1}{c|}{\textbf{P}} & \multicolumn{1}{c|}{\textbf{R}} & \textbf{F1} & \multicolumn{1}{c|}{\textbf{P}} & \multicolumn{1}{c|}{\textbf{R}} & \textbf{F1} & \multicolumn{1}{c|}{\textbf{P}} & \multicolumn{1}{c|}{\textbf{R}} & \textbf{F1} & \multicolumn{1}{c|}{\textbf{P}} & \multicolumn{1}{c|}{\textbf{R}} & \textbf{F1} & \multicolumn{1}{c|}{\textbf{P}} & \multicolumn{1}{c}{\textbf{R}} & \textbf{F1} \\ \hline
\textbf{Gemma (0-shot)} & \multicolumn{1}{c|}{12.3} & \multicolumn{1}{c|}{13.0} & 12.6 & \multicolumn{1}{c|}{27.3} & \multicolumn{1}{c|}{27.3} & 27.3 & \multicolumn{1}{c|}{1.6} & \multicolumn{1}{c|}{1.6} & 1.6 & \multicolumn{1}{c|}{4.8} & \multicolumn{1}{c|}{6.4} & 5.5 & \multicolumn{1}{c|}{3.3} & \multicolumn{1}{c|}{5.3} & 4.1 \\ 
\textbf{Gemma (5-shot)} & \multicolumn{1}{c|}{27.2} & \multicolumn{1}{c|}{28.4} & 27.8 & \multicolumn{1}{c|}{51.6} & \multicolumn{1}{c|}{51.2} & 51.4 & \multicolumn{1}{c|}{14.8} & \multicolumn{1}{c|}{14.8} & 14.8 & \multicolumn{1}{c|}{32.3} & \multicolumn{1}{c|}{41.9} & \textbf{36.5} & \multicolumn{1}{c|}{7.4} & \multicolumn{1}{c|}{13.0} & 9.4 \\ 
\textbf{Llama 8B (0-shot)} & \multicolumn{1}{c|}{25.8} & \multicolumn{1}{c|}{27.3} & 26.5 & \multicolumn{1}{c|}{31.9} & \multicolumn{1}{c|}{27.0} & 29.2 & \multicolumn{1}{c|}{2.4} & \multicolumn{1}{c|}{2.4} & 2.4 & \multicolumn{1}{c|}{9.2} & \multicolumn{1}{c|}{11.8} & 10.3 & \multicolumn{1}{c|}{3.9} & \multicolumn{1}{c|}{5.1} & 4.4 \\ 
\textbf{Llama 8B (5-shot)} & \multicolumn{1}{c|}{33.8} & \multicolumn{1}{c|}{35.5} & 34.6 & \multicolumn{1}{c|}{55.3} & \multicolumn{1}{c|}{54.0} & 54.6 & \multicolumn{1}{c|}{11.5} & \multicolumn{1}{c|}{11.5} & 11.5 & \multicolumn{1}{c|}{22.1} & \multicolumn{1}{c|}{28.8} & 25.0 & \multicolumn{1}{c|}{10.4} & \multicolumn{1}{c|}{12.3} & \textbf{11.3} \\ 
\textbf{Llama 70B (0-shot)} & \multicolumn{1}{c|}{27.2} & \multicolumn{1}{c|}{28.6} & 27.9 & \multicolumn{1}{c|}{41.0} & \multicolumn{1}{c|}{40.6} & 40.8 & \multicolumn{1}{c|}{9.2} & \multicolumn{1}{c|}{9.2} & 9.2 & \multicolumn{1}{c|}{17.9} & \multicolumn{1}{c|}{23.8} & 20.4 & \multicolumn{1}{c|}{9.3} & \multicolumn{1}{c|}{8.8} & 9.0 \\ 
\textbf{Llama 70B (5-shot)} & \multicolumn{1}{c|}{30.5} & \multicolumn{1}{c|}{32.0} & 31.2 & \multicolumn{1}{c|}{64.2} & \multicolumn{1}{c|}{62.5} & \textbf{63.3} & \multicolumn{1}{c|}{15.3} & \multicolumn{1}{c|}{15.3} & \textbf{15.3} & \multicolumn{1}{c|}{23.5} & \multicolumn{1}{c|}{30.9} & 26.7 & \multicolumn{1}{c|}{8.5} & \multicolumn{1}{c|}{8.4} & 8.4 \\ 
\textbf{Mistral (0-shot)} & \multicolumn{1}{c|}{14.3} & \multicolumn{1}{c|}{15.0} & 14.6 & \multicolumn{1}{c|}{25.8} & \multicolumn{1}{c|}{24.3} & 25.0 & \multicolumn{1}{c|}{1.4} & \multicolumn{1}{c|}{1.4} & 1.4 & \multicolumn{1}{c|}{6.5} & \multicolumn{1}{c|}{8.6} & 7.4 & \multicolumn{1}{c|}{0.2} & \multicolumn{1}{c|}{0.4} & 0.3 \\ 
\textbf{Mistral (5-shot)} & \multicolumn{1}{c|}{34.8} & \multicolumn{1}{c|}{36.6} & \textbf{35.7} & \multicolumn{1}{c|}{53.0} & \multicolumn{1}{c|}{51.0} & 52.0 & \multicolumn{1}{c|}{15.2} & \multicolumn{1}{c|}{15.2} & 15.2 & \multicolumn{1}{c|}{22.5} & \multicolumn{1}{c|}{30} & 25.7 & \multicolumn{1}{c|}{7.8} & \multicolumn{1}{c|}{13.0} & 9.7 \\ 
\textbf{Qwen 4B (0-shot)} & \multicolumn{1}{c|}{26.1} & \multicolumn{1}{c|}{27.6} & 26.8 & \multicolumn{1}{c|}{31.5} & \multicolumn{1}{c|}{31.0} & 31.2 & \multicolumn{1}{c|}{0.5} & \multicolumn{1}{c|}{0.5} & 0.5 & \multicolumn{1}{c|}{3.2} & \multicolumn{1}{c|}{4.3} & 3.7 & \multicolumn{1}{c|}{0.8} & \multicolumn{1}{c|}{1.8} & 1.1 \\ 
\textbf{Qwen 4B (5-shot} & \multicolumn{1}{c|}{34.8} & \multicolumn{1}{c|}{36.4} & 35.6 & \multicolumn{1}{c|}{59.6} & \multicolumn{1}{c|}{56.9} & 58.2 & \multicolumn{1}{c|}{7.4} & \multicolumn{1}{c|}{7.4} & 7.4 & \multicolumn{1}{c|}{21.2} & \multicolumn{1}{c|}{27.0} & 23.8 & \multicolumn{1}{c|}{4.6} & \multicolumn{1}{c|}{7.9} & 5.8 \\ 
\textbf{Qwen 32B (0-shot)} & \multicolumn{1}{c|}{19.5} & \multicolumn{1}{c|}{18.2} & 18.8 & \multicolumn{1}{c|}{34.1} & \multicolumn{1}{c|}{27.5} & 30.4 & \multicolumn{1}{c|}{1.4} & \multicolumn{1}{c|}{1.2} & 1.3 & \multicolumn{1}{c|}{7.7} & \multicolumn{1}{c|}{8.6} & 8.1 & \multicolumn{1}{c|}{1.3} & \multicolumn{1}{c|}{1.8} & 1.5 \\ 
\textbf{Qwen 32B (5-shot)} & \multicolumn{1}{c|}{29.0} & \multicolumn{1}{c|}{24.5} & 26.6 & \multicolumn{1}{c|}{46.4} & \multicolumn{1}{c|}{34.6} & 39.6 & \multicolumn{1}{c|}{8.8} & \multicolumn{1}{c|}{6.8} & 7.7 & \multicolumn{1}{c|}{25.1} & \multicolumn{1}{c|}{22.6} & 23.8 & \multicolumn{1}{c|}{4.8} & \multicolumn{1}{c|}{5.3} & 5.0 \\ 
\textbf{T5 Large (0-shot)} & \multicolumn{1}{c|}{0.0} & \multicolumn{1}{c|}{0.0} & 0.0 & \multicolumn{1}{c|}{0.0} & \multicolumn{1}{c|}{0.0} & 0.0 & \multicolumn{1}{c|}{0.0} & \multicolumn{1}{c|}{0.0} & 0.0 & \multicolumn{1}{c|}{0.0} & \multicolumn{1}{c|}{0.0} & 0.0 & \multicolumn{1}{c|}{0.0} & \multicolumn{1}{c|}{0.0} & 0.0 \\ 
\textbf{T5 Large (5-shot)} & \multicolumn{1}{c|}{1.2} & \multicolumn{1}{c|}{1.3} & 1.2 & \multicolumn{1}{c|}{1.9} & \multicolumn{1}{c|}{2} & 1.9 & \multicolumn{1}{c|}{0.0} & \multicolumn{1}{c|}{0.0} & 0.0 & \multicolumn{1}{c|}{0.1} & \multicolumn{1}{c|}{0.1} & 0.1 & \multicolumn{1}{c|}{0.1} & \multicolumn{1}{c|}{0.2} & 0.1 \\ \hline
\end{tabular}}
\caption{Performance comparison on event extraction (\%) using EM. Bold indicates the best performance.}
\label{table6}
\end{table*}

% \usepackage{multirow}
\begin{table*}[]
\centering
\resizebox{\textwidth}{!}{
\begin{tabular}{l|ccc|ccc|ccc|ccc|ccc}
\hline
\multicolumn{1}{c|}{\multirow{2}{*}{\textbf{Model}}} & \multicolumn{3}{c|}{\textbf{Where}} & \multicolumn{3}{c|}{\textbf{When}} & \multicolumn{3}{c|}{\textbf{What}} & \multicolumn{3}{c|}{\textbf{Who}} & \multicolumn{3}{c}{\textbf{Why}} \\ \cline{2-16} 
\multicolumn{1}{c|}{} & \multicolumn{1}{c|}{\textbf{P}} & \multicolumn{1}{c|}{\textbf{R}} & \textbf{F1} & \multicolumn{1}{c|}{\textbf{P}} & \multicolumn{1}{c|}{\textbf{R}} & \textbf{F1} & \multicolumn{1}{c|}{\textbf{P}} & \multicolumn{1}{c|}{\textbf{R}} & \textbf{F1} & \multicolumn{1}{c|}{\textbf{P}} & \multicolumn{1}{c|}{\textbf{R}} & \textbf{F1} & \multicolumn{1}{c|}{\textbf{P}} & \multicolumn{1}{c}{\textbf{R}} & \textbf{F1} \\ \hline
\textbf{Gemma (0-shot)} & \multicolumn{1}{c|}{34.3} & \multicolumn{1}{c|}{78.9} & 43.4 & \multicolumn{1}{c|}{42.4} & \multicolumn{1}{c|}{73.3} & 48.9 & \multicolumn{1}{c|}{28.6} & \multicolumn{1}{c|}{51.8} & 32.0 & \multicolumn{1}{c|}{22.9} & \multicolumn{1}{c|}{65.3} & 30.3 & \multicolumn{1}{c|}{31.4} & \multicolumn{1}{c|}{53.6} & 36.0 \\ 
\textbf{Gemma (5-shot)} & \multicolumn{1}{c|}{44.3} & \multicolumn{1}{c|}{64.7} & 48.3 & \multicolumn{1}{c|}{64.3} & \multicolumn{1}{c|}{78.3} & 68.1 & \multicolumn{1}{c|}{52.4} & \multicolumn{1}{c|}{57.7} & \textbf{49.2} & \multicolumn{1}{c|}{59.9} & \multicolumn{1}{c|}{75.6} & \textbf{63.4} & \multicolumn{1}{c|}{46.6} & \multicolumn{1}{c|}{59.0} & 48.4 \\ 
\textbf{Llama 8B (0-shot)} & \multicolumn{1}{c|}{39.2} & \multicolumn{1}{c|}{54.7} & 42.0 & \multicolumn{1}{c|}{44.0} & \multicolumn{1}{c|}{65.4} & 48.5 & \multicolumn{1}{c|}{30.9} & \multicolumn{1}{c|}{41.5} & 29.4 & \multicolumn{1}{c|}{26.7} & \multicolumn{1}{c|}{56.7} & 32.5 & \multicolumn{1}{c|}{33.9} & \multicolumn{1}{c|}{42.5} & 34.4 \\ 
\textbf{Llama 8B (5-shot)} & \multicolumn{1}{c|}{47.8} & \multicolumn{1}{c|}{62.6} & \textbf{50.6} & \multicolumn{1}{c|}{66.7} & \multicolumn{1}{c|}{76.9} & 69.8 & \multicolumn{1}{c|}{50.6} & \multicolumn{1}{c|}{51.6} & 44.8 & \multicolumn{1}{c|}{43.3} & \multicolumn{1}{c|}{62.5} & 47.3 & \multicolumn{1}{c|}{52.4} & \multicolumn{1}{c|}{61.0} & 52.3 \\ 
\textbf{Llama 70B (0-shot)} & \multicolumn{1}{c|}{41.4} & \multicolumn{1}{c|}{60.6} & 45.1 & \multicolumn{1}{c|}{52.6} & \multicolumn{1}{c|}{64.1} & 56.0 & \multicolumn{1}{c|}{44.1} & \multicolumn{1}{c|}{41.8} & 36.6 & \multicolumn{1}{c|}{36.3} & \multicolumn{1}{c|}{49.5} & 39.1 & \multicolumn{1}{c|}{45.7} & \multicolumn{1}{c|}{52.9} & 45.9 \\ 
\textbf{Llama 70B (5-shot)} & \multicolumn{1}{c|}{44.9} & \multicolumn{1}{c|}{64.6} & 49.0 & \multicolumn{1}{c|}{73.2} & \multicolumn{1}{c|}{79.5} & \textbf{75.2} & \multicolumn{1}{c|}{51.0} & \multicolumn{1}{c|}{47.2} & 43.8 & \multicolumn{1}{c|}{43.7} & \multicolumn{1}{c|}{55.2} & 46.4 & \multicolumn{1}{c|}{51.5} & \multicolumn{1}{c|}{63.6} & \textbf{53.5} \\ 
\textbf{Mistral (0-shot)} & \multicolumn{1}{c|}{38.2} & \multicolumn{1}{c|}{75.8} & 46.9 & \multicolumn{1}{c|}{38.4} & \multicolumn{1}{c|}{65.9} & 43.8 & \multicolumn{1}{c|}{27.0} & \multicolumn{1}{c|}{62.2} & 32.9 & \multicolumn{1}{c|}{23.1} & \multicolumn{1}{c|}{63.5} & 29.9 & \multicolumn{1}{c|}{15.9} & \multicolumn{1}{c|}{47.1} & 21.3 \\ 
\textbf{Mistral (5-shot)} & \multicolumn{1}{c|}{47.0} & \multicolumn{1}{c|}{59.6} & 49.0 & \multicolumn{1}{c|}{63.1} & \multicolumn{1}{c|}{72.0} & 65.8 & \multicolumn{1}{c|}{56.7} & \multicolumn{1}{c|}{45.6} & 44.8 & \multicolumn{1}{c|}{44.0} & \multicolumn{1}{c|}{62.9} & 47.7 & \multicolumn{1}{c|}{46.4} & \multicolumn{1}{c|}{58.2} & 47.6 \\ 
\textbf{Qwen 4B (0-shot)} & \multicolumn{1}{c|}{42.0} & \multicolumn{1}{c|}{64.2} & 46.0 & \multicolumn{1}{c|}{46.1} & \multicolumn{1}{c|}{71.8} & 51.8 & \multicolumn{1}{c|}{23.6} & \multicolumn{1}{c|}{61.4} & 30.2 & \multicolumn{1}{c|}{19.6} & \multicolumn{1}{c|}{65.6} & 27.2 & \multicolumn{1}{c|}{21.2} & \multicolumn{1}{c|}{48.7} & 26.7 \\ 
\textbf{Qwen 4B (5-shot} & \multicolumn{1}{c|}{44.3} & \multicolumn{1}{c|}{52.5} & 45.4 & \multicolumn{1}{c|}{68.6} & \multicolumn{1}{c|}{77.1} & 71.0 & \multicolumn{1}{c|}{37.5} & \multicolumn{1}{c|}{63.9} & 42.4 & \multicolumn{1}{c|}{41.4} & \multicolumn{1}{c|}{62.2} & 45.5 & \multicolumn{1}{c|}{35.7} & \multicolumn{1}{c|}{56.4} & 39.4 \\ 
\textbf{Qwen 32B (0-shot)} & \multicolumn{1}{c|}{40.0} & \multicolumn{1}{c|}{77.5} & 47.6 & \multicolumn{1}{c|}{47.2} & \multicolumn{1}{c|}{70.9} & 52.5 & \multicolumn{1}{c|}{22.7} & \multicolumn{1}{c|}{59.5} & 29.0 & \multicolumn{1}{c|}{27.8} & \multicolumn{1}{c|}{71.1} & 35.4 & \multicolumn{1}{c|}{24.9} & \multicolumn{1}{c|}{48.2} & 29.5 \\ 
\textbf{Qwen 32B (5-shot)} & \multicolumn{1}{c|}{43.9} & \multicolumn{1}{c|}{66.7} & 48.3 & \multicolumn{1}{c|}{57.1} & \multicolumn{1}{c|}{76.1} & 60.7 & \multicolumn{1}{c|}{38.2} & \multicolumn{1}{c|}{61.8} & 41.9 & \multicolumn{1}{c|}{43.8} & \multicolumn{1}{c|}{61.5} & 46.5 & \multicolumn{1}{c|}{38.9} & \multicolumn{1}{c|}{52.8} & 40.1 \\ 
\textbf{T5 Large (0-shot)} & \multicolumn{1}{c|}{0.0} & \multicolumn{1}{c|}{0.0} & 0.0 & \multicolumn{1}{c|}{0.0} & \multicolumn{1}{c|}{0.0} & 0.0 & \multicolumn{1}{c|}{0.0} & \multicolumn{1}{c|}{0.0} & 0.0 & \multicolumn{1}{c|}{0.0} & \multicolumn{1}{c|}{0.0} & 0.0 & \multicolumn{1}{c|}{0.0} & \multicolumn{1}{c|}{0.0} & 0.0 \\ 
\textbf{T5 Large (5-shot)} & \multicolumn{1}{c|}{1.4} & \multicolumn{1}{c|}{1.3} & 1.3 & \multicolumn{1}{c|}{2.0} & \multicolumn{1}{c|}{2.0} & 2.0 & \multicolumn{1}{c|}{0.8} & \multicolumn{1}{c|}{0.3} & 0.4 & \multicolumn{1}{c|}{0.3} & \multicolumn{1}{c|}{0.2} & 0.2 & \multicolumn{1}{c|}{0.7} & \multicolumn{1}{c|}{0.5} & 0.4 \\ \hline
\end{tabular}}
\caption{Performance comparison on event extraction (\%) using ROGUE-L. Bold indicates the best performance.}
\label{table7}
\end{table*}

\section{Generality of Algorithms Trained on EVENT5Ws}
To assess the generality of algorithms trained on EVENT5Ws, we first fine-tune T5 Large, the weakest-performing model in Section 5.3, on the training and validation sets. The model is trained for 10 epochs using a learning rate of 1e-4, weight decay of 1e-2, batch size of 8, and beam size of 5 on NVIDIA V100 GPUs. We then evaluate the trained model on a manually verified main event 5Ws extraction dataset \cite{hamborgGiveme5W1HUniversalSystem2019}, which consists of 96 documents from newspapers in the USA and the UK. We also compare its performance with Givme5W1H  \cite{hamborgGiveme5W1HUniversalSystem2019}, an open-source main event 5Ws event extraction algorithm developed using data from similar contexts. 

Tables \ref{table8} and \ref{table9} present the evaluation results using precision (P), recall (R), and F1 score computed using EM and ROGUE-L. As shown in the results, the fine-tuned T5 Large outperforms the Giveme5W1H across all Ws for both EM and ROGUE-L. The lower performance of Giveme5W1H may be attributed to its reliance on rules and heuristics, which limits its ability to generalize beyond predefined rules and linguistic patterns. Despite being trained on EVENT5Ws, which contains documents from a different geographical and cultural context (i.e., India), the T5 Large generalizes effectively to data from the US and UK. This highlights the potential of EVENT5Ws for training robust event extraction approaches that generalize across diverse contexts.

% \usepackage{multirow}
\begin{table*}[]
\centering
\resizebox{\textwidth}{!}{
\begin{tabular}{l|ccc|ccc|ccc|ccc|ccc}
\hline
\multicolumn{1}{c|}{\multirow{2}{*}{\textbf{Model}}} & \multicolumn{3}{c|}{\textbf{Where}} & \multicolumn{3}{c|}{\textbf{When}} & \multicolumn{3}{c|}{\textbf{What}} & \multicolumn{3}{c|}{\textbf{Who}} & \multicolumn{3}{c}{\textbf{Why}} \\ \cline{2-16} 
\multicolumn{1}{c|}{} & \multicolumn{1}{c|}{\textbf{P}} & \multicolumn{1}{c|}{\textbf{R}} & \textbf{F1} & \multicolumn{1}{c|}{\textbf{P}} & \multicolumn{1}{c|}{\textbf{R}} & \textbf{F1} & \multicolumn{1}{c|}{\textbf{P}} & \multicolumn{1}{c|}{\textbf{R}} & \textbf{F1} & \multicolumn{1}{c|}{\textbf{P}} & \multicolumn{1}{c|}{\textbf{R}} & \textbf{F1} & \multicolumn{1}{c|}{\textbf{P}} & \multicolumn{1}{c}{\textbf{R}} & \textbf{F1} \\ \hline
\textbf{T5 Large} & \multicolumn{1}{c|}{46.3} & \multicolumn{1}{c|}{33.7} & \textbf{39.0} & \multicolumn{1}{c|}{29.2} & \multicolumn{1}{c|}{28.9} & \textbf{29.0} & \multicolumn{1}{c|}{27.1} & \multicolumn{1}{c|}{19.3} & \textbf{22.5} & \multicolumn{1}{c|}{70.7} & \multicolumn{1}{c|}{33.8} & \textbf{45.7} & \multicolumn{1}{c|}{10.7} & \multicolumn{1}{c|}{2.6} & \textbf{4.2} \\ 
\textbf{Givme5W1H} & \multicolumn{1}{c|}{20.5} & \multicolumn{1}{c|}{16.3} & 18.2 & \multicolumn{1}{c|}{30.2} & \multicolumn{1}{c|}{26.8} & 28.4 & \multicolumn{1}{c|}{6.2} & \multicolumn{1}{c|}{4.4} & 5.1 & \multicolumn{1}{c|}{30.2} & \multicolumn{1}{c|}{18.5} & 22.9 & \multicolumn{1}{c|}{0.0} & \multicolumn{1}{c|}{0.0} & 0.0 \\ \hline
\end{tabular}}
\caption{Performance comparison on event extraction (\%) using EM for T5 Large and Giveme5W1H. Bold indicates the best performance.}
\label{table8}
\end{table*}

% \usepackage{multirow}
\begin{table*}[]
\centering
\resizebox{\textwidth}{!}{
\begin{tabular}{l|ccc|ccc|ccc|ccc|ccc}
\hline
\multicolumn{1}{c|}{\multirow{2}{*}{\textbf{Model}}} & \multicolumn{3}{c|}{\textbf{Where}} & \multicolumn{3}{c|}{\textbf{When}} & \multicolumn{3}{c|}{\textbf{What}} & \multicolumn{3}{c|}{\textbf{Who}} & \multicolumn{3}{c}{\textbf{Why}} \\ \cline{2-16} 
\multicolumn{1}{c|}{} & \multicolumn{1}{c|}{\textbf{P}} & \multicolumn{1}{c|}{\textbf{R}} & \textbf{F1} & \multicolumn{1}{c|}{\textbf{P}} & \multicolumn{1}{c|}{\textbf{R}} & \textbf{F1} & \multicolumn{1}{c|}{\textbf{P}} & \multicolumn{1}{c|}{\textbf{R}} & \textbf{F1} & \multicolumn{1}{c|}{\textbf{P}} & \multicolumn{1}{c|}{\textbf{R}} & \textbf{F1} & \multicolumn{1}{c|}{\textbf{P}} & \multicolumn{1}{c|}{\textbf{R}} & \textbf{F1} \\ \hline
\textbf{T5 Large} & \multicolumn{1}{c|}{75.8} & \multicolumn{1}{c|}{58.5} & \textbf{62.2} & \multicolumn{1}{c|}{56.8} & \multicolumn{1}{c|}{43} & \textbf{46.5} & \multicolumn{1}{c|}{62.7} & \multicolumn{1}{c|}{70.2} & \textbf{61.5} & \multicolumn{1}{c|}{83.8} & \multicolumn{1}{c|}{83.1} & \textbf{81.9} & \multicolumn{1}{c|}{47.7} & \multicolumn{1}{c|}{46.3} & \textbf{43.9} \\ 
\textbf{Givme5W1H} & \multicolumn{1}{c|}{45.6} & \multicolumn{1}{c|}{34.3} & 37.1 & \multicolumn{1}{c|}{40.7} & \multicolumn{1}{c|}{41.6} & 40.9 & \multicolumn{1}{c|}{33.1} & \multicolumn{1}{c|}{38.5} & 32.2 & \multicolumn{1}{c|}{52.5} & \multicolumn{1}{c|}{63.7} & 52.5 & \multicolumn{1}{c|}{5.6} & \multicolumn{1}{c|}{9.4} & 6.1 \\ \hline
\end{tabular}}
\caption{Performance comparison on event extraction (\%) using ROGUE-L for T5 Large and Giveme5W1H. Bold indicates the best performance.}
\label{table9}
\end{table*}
\section{Lessons Learned from 5Ws Annotation and Recommendations}
The annotation of main event 5Ws provided several insights into creating large-scale datasets involving multiple coders and multiple labels. In this section, we summarize key lessons learned during the development of EVENT5Ws and provide recommendations for future efforts on similar annotation tasks.

\textbf{Coder Recruitment: }Coders play a critical role in ensuring annotation quality and thus selecting suitable coders is vital for the success of any annotation task. Coders with domain knowledge (i.e., linguistics, social, and cultural familiarity with dataset context) are generally more efficient and consistent. The high ICR values during coder training (see Appendix \ref{appen4}) support this as the coders were well-versed with the geographic scope of our dataset.

Having domain knowledge also improves the annotation time. To evaluate this, we conducted an experiment in which our coders annotated Where and When for 150 documents from Mexican newspapers (a context unfamiliar to them) and 150 documents from Indian newspapers not included in EVENT5Ws. Here, we did not use the annotations from the EVENT5Ws due to the large difference in size (10,000 in EVENT5Ws to 150 from the Mexican newspapers), which may affect the validity of the results. The average annotation time was 43.73 seconds for Indian documents and 68.59 seconds for Mexican documents. Additionally, a Kruskal-Wallis\footnote{We applied the Kruskal-Wallis test due to violations of the normality assumption in our data.} test \cite{kruskalUseRanksOneCriterion1952} on the annotation times for the two contexts resulted in a p-value of 0.047, which is below the significance threshold of 0.05 and indicates a statistically significant difference in annotation time distributions. This difference in distribution, along with the higher average annotation time, suggests that our coders required more time to annotate documents from Mexico than India. Based on these findings, we recommend recruiting coders familiar with the dataset context.

\textbf{Annotation Process: }The design of the annotation process significantly impacts coder efficiency. We adopted a sequencing strategy, where different Ws were annotated in separate phases. This strategy reduced monotony as coders could focus on different Ws at different phases. In addition, it allowed us to replace a coder when necessary and gave continuity to dataset development. While the strategy provided flexibility, it required coders to read the same set of documents multiple times across different phases. This increased the annotation time and cost. However, annotating a large corpus is a long process, and retaining coders for the entire duration is often challenging. This is especially difficult when the coders are university students, as they may graduate or have other academic commitments. Based on our experience, the benefits of the sequencing strategy outweigh the additional cost and time involved, as it was crucial in the successful development of EVENT5Ws.

Another strategy we employed was annotating the dataset in smaller batches. This strategy helped in maintaining coder engagement and reducing fatigue. In addition, it allowed us to regularly discuss the disagreements among coders. Given the benefits discussed above, we suggest using a sequencing strategy and dividing dataset into different batches, specifically when the annotation involves a large dataset, multiple coders, and multiple labels.

\textbf{Annotation and Resolution Guidelines: } Clear and well-defined guidelines streamline the annotation and resolution process. Therefore, we suggest developing easy-to-understand guidelines. In our study, we gave clear guidelines along with examples to the coders. This helped coders better understand the annotation task as showcased by the high ICR values during training. In addition, we provided access to previous annotations during the resolution step. This allowed coders to have contextual information about previous annotations, which proved beneficial as coders became more efficient in disagreement resolution as demonstrated by the reduced annotation time during the resolution step (see Table \ref{table4}) and the low number of documents the experts had to resolve (see Section 4.4). As a result, we recommend providing coders with clear guidelines that include examples of annotations and access to their previous annotations during disagreement resolution.

\textbf{Planning: }Creating large, manually annotated datasets is time-consuming and costly. Therefore, good planning is necessary for successful completion. Based on our experience, it is crucial to spend time finding an appropriate annotation platform, selecting coders familiar with dataset context, and designing an annotation pipeline tailored to the annotation task.

\section{Conclusion}In this work, we introduced EVENT5Ws, a large-scale, manually annotated dataset for open-domain event extraction based on the 5Ws framework. We described the annotation process, including the use of university students as coders and validation through inter-coder reliability, and provided insights into large-scale dataset development. Through extensive experiments, we established baselines using SOTA LLMs for open-domain event extraction and also demonstrated the challenging nature of document-level event extraction. Furthermore, we showed that models trained on EVENT5Ws generalize effectively to datasets from diverse geographical and textual contexts and highlighted the potential of EVENT5Ws for developing generalizable event extraction algorithms. Moving forward, we plan to incorporate additional multimedia sources, such as newspaper images, to enrich the dataset and support multimodal event extraction.

\bibliography{tacl2021}
\bibliographystyle{acl_natbib}

\appendix 
\section{Annotation Platform and Dataset Details} \label{appen1}

\begin{figure*}[t]
\centering
\includegraphics[width=\linewidth]{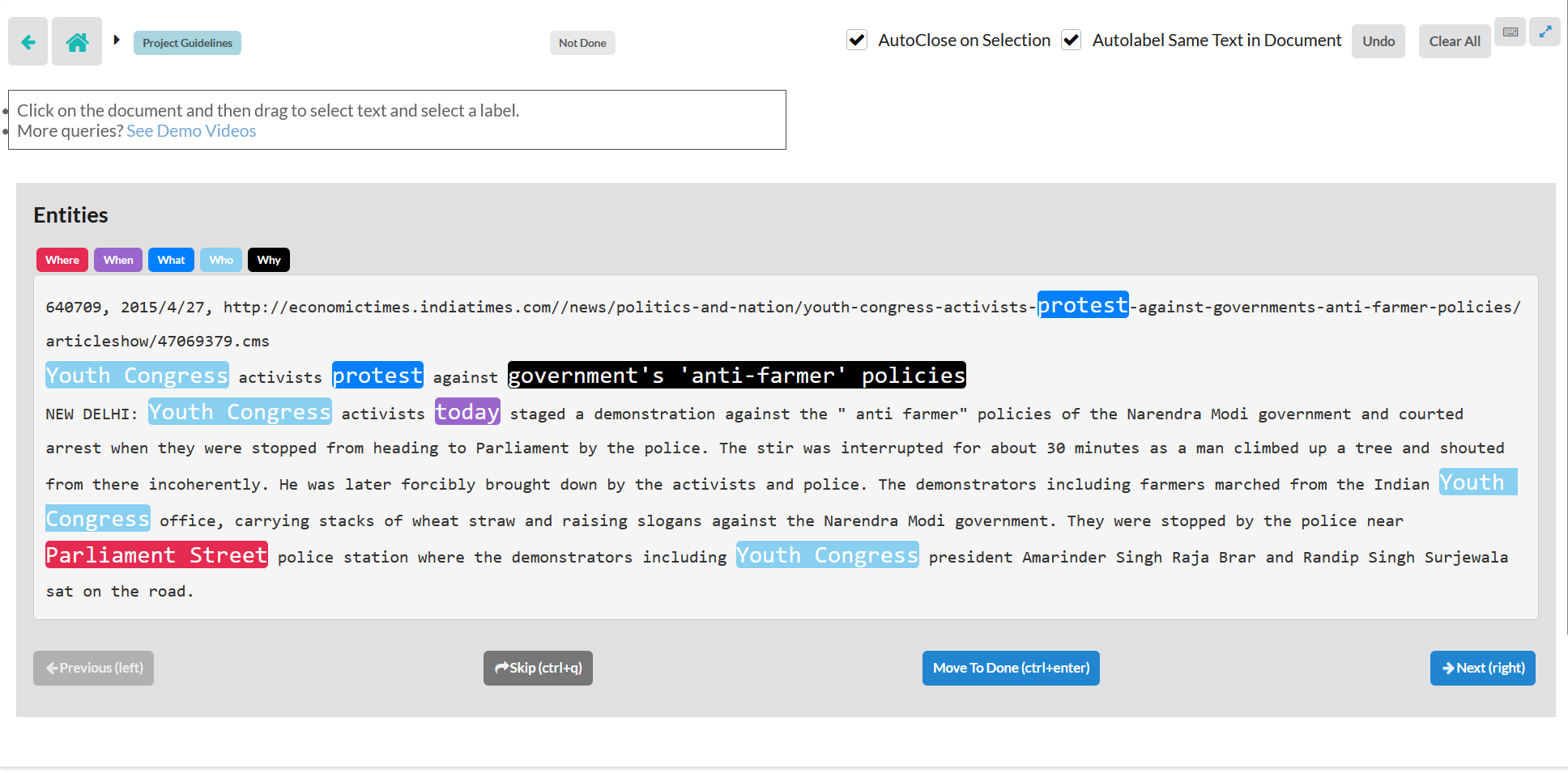}
\caption{Snapshot of Dataturks, our annotation platform, and an example of 5Ws annotated for the main event in a document from our dataset.}
\label{figure2}
\end{figure*}

We use Dataturks as our annotation platform for annotating the 5Ws. The platform provides an intuitive interface that allows coders to highlight text spans corresponding to each W and assign labels efficiently. It also supports multi-user annotation, enabling multiple coders to work concurrently and facilitating the management of large-scale annotation tasks. Figure~\ref{figure2} shows a snapshot of the annotation interface and an example document from our dataset with 5Ws for the main event annotated (highlighted).

\begin{figure*}[t]
\centering
\includegraphics[width=\linewidth]{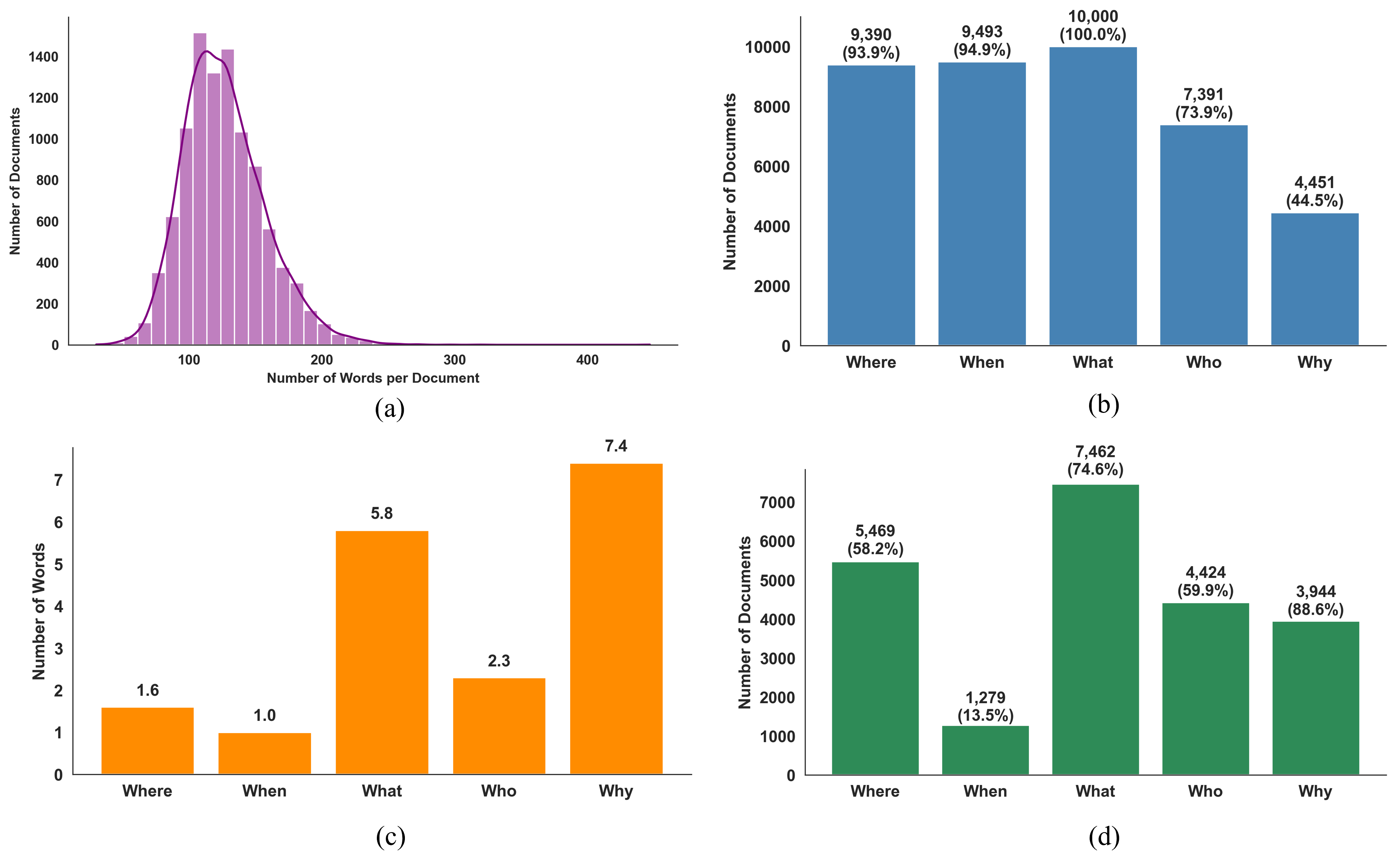}
\caption{Summary of the EVENT5Ws dataset: (a) document length distribution, (b) number of documents with annotations for each W, (c) span length (in words) across the 5Ws, and (d) number of documents with unique values for each W.}
\label{datasetsummary}
\end{figure*}

Figure \ref{datasetsummary} summarizes key properties of the dataset. Figure 3a shows the distribution of the document lengths (number of words), which are concentrated around 100-150 words with a long tail toward longer documents. On average, the documents contain 127 words. Figure 3b shows the number of documents containing annotations for each W. What is present in all documents, while Where (93.9\%), When (94.9\%), and Who (73.9\%) are also frequent. In contrast, Why is comparatively sparse and present in only 44.5\% of documents. This reflects the limited availability of causal information and highlights the difficulty of extracting Why. 

The documents in the dataset contain multiple annotated Ws, averaging 4.1 per document. Figure 3c highlights differences in span length (number of words) across the 5Ws. Where, When, and Who are generally short, whereas What and Why tend to be longer, which reflect their descriptive nature. These variations suggest differing levels of extraction complexity across the Ws. Figure 3d demonstrates the lexical diversity of the dataset. What (75.2\%) and Why (88.9\%) exhibit high diversity, while When (13.6\%) remains comparatively limited. This diversity indicates limited repetition and increased difficulty for generalization.

\section{Annotation Guidelines} \label{appen2}
\subsection{What Annotation}
\begin{figure*}[t]
\centering
\includegraphics[width=\linewidth]{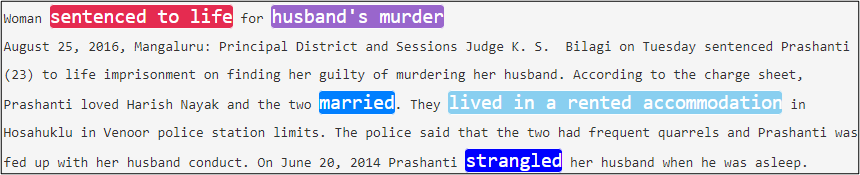}
\caption{Illustration of the main event described in a document. The main event is highlighted in red and all background events are highlighted in other colors.}
\label{figure3}
\end{figure*}

\begin{figure*}[t]
\centering
\includegraphics[width=\linewidth]{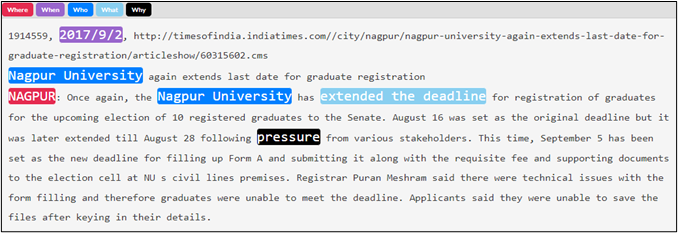}
\caption{An example of What annotation in a structured document highlighted in light blue.}
\label{figure4}
\end{figure*}

What represents the concept of the main event described in a document. It is typically expressed as event nuggets (i.e., semantically meaningful units), which can be a single word (verb, noun, or adjective) or a phrase. What Annotation involves identifying and marking the nugget that best describes the main event. The guidelines for What annotation are:
\begin{itemize}
    \item Identify the most recent and newsworthy event described in a document and annotate the event nugget that conveys its concept. For example, in Figure \ref{figure3}, the most recent and newsworthy event described in the document is ‘sentenced to life’ and should be annotated. However, events such as ‘husband’s murder’ and ‘married’ are background (or precursor) events and should not be annotated.
    \item Annotate the most concise and meaningful event nugget that best describes the concept of the reported main event. For example, in Figure \ref{figure4}, the event nugget ‘extended the deadline’ mentioned in the document most concisely indicates the concept of the main event described and should be annotated. Choosing a shorter event nugget, e.g., ‘extended,’ ‘deadline’, would not sufficiently provide complete information on the main event.
\end{itemize}

\subsection{Where Annotation}
\begin{figure*}[t]
\centering
\includegraphics[width=\linewidth]{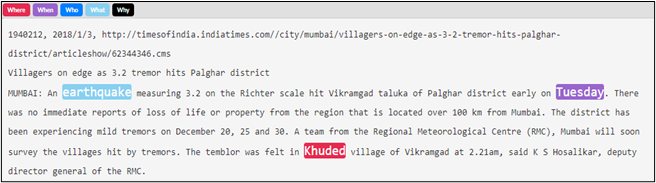}
\caption{An example of Where annotation (highlighted in red) for the main event ‘earthquake.’ ‘Khuded’ is the most precise location of occurrence of the main event ‘earthquake’. Note that there are other place names that appear in the document: Vikramgad, Palghar, and Mumbai.}
\label{figure5}
\end{figure*}

Where represents the location of the main event described in a document. Annotating Where involves identifying and marking the place where the main event occurred. The guidelines are:
\begin{itemize}
    \item Identify and annotate the place name (location) with the highest spatial resolution mentioned in a document where the reported main event occurred (see Figure \ref{figure5}).
    \item If a document does not contain any information about the location of the main event, no annotation is required.
\end{itemize}

\subsection{When Annotation}
When represents the date or time of occurrence of the main event described in a document. Annotating When involves identifying and marking the temporal information that specifies when the main event occurred. The guidelines are:
\begin{itemize}
    \item Identify and annotate the date or time of occurrence of the main event described in a document. This can appear as a day (e.g., Sunday, Monday, January 1, October 10), month (e.g., January, February), year (e.g., 1992, 2022), or full date (e.g., January 1, 2022), as well as temporal adverbs or nouns (e.g., today, yesterday). For example, in Figure \ref{figure5}, the When for the main event 'earthquake' is 'Tuesday', highlighted in purple.
    \item If a news report does not contain any information about the date or time of the main event, no annotation is required.
\end{itemize}

\subsection{Who Annotation}
\begin{figure*}[t]
\centering
\includegraphics[width=\linewidth]{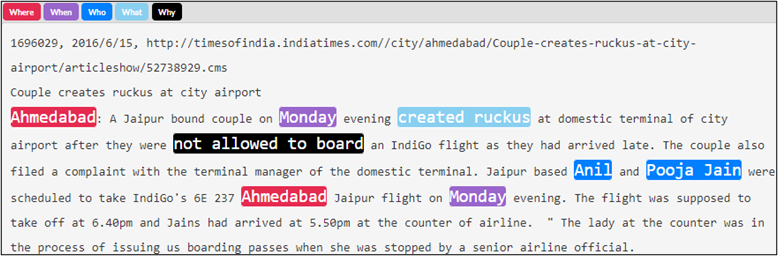}
\caption{An example of Who annotation highlighted in dark blue for the main event ‘created ruckus.’ Two people are the actors for the event, and thus both are annotated. Note that in this study, the annotation of specific entities is preferred over the generic ones for the Who when both are mentioned, as they offer more precision. As a result, ‘Anil’ and ‘Pooja Jain’ are annotated as ‘who’ instead of ‘couple.’}
\label{figure6}
\end{figure*}

Who represents the primary actor or instigator responsible for the main event described in a document. Annotating Who involves identifying and marking the entity or entities that participate in or are responsible for the main event. The guidelines are:
\begin{itemize}
    \item Identify and annotate the primary actor (person(s) or organization(s)) that play a central role in the main event described in the document (see Figure \ref{figure6}).
    \item In some cases, a specific person or an organization may not be the primary actor of the main event. Instead, a generic group (e.g., students, politicians, doctors, etc.) may be responsible for the event. In such cases, identify and annotate the generic group.
    \item If a document does not contain any information about the primary actor(s) of the main event, no annotation is required.
\end{itemize}

\subsection{Why Annotation}
Why represents the cause or reason of the occurrence of the main event described in a document. Similar to the What, annotating Why involves identifying and marking a semantically meaningful unit, i.e., a single word (verb, noun, or adjective) or a phrase, that best captures the cause of the main event. The guidelines are:
\begin{itemize}
    \item Identify and annotate the most concise phrase or word that best describes the cause of the main event. For example, in Figure \ref{figure6}, the phrase ‘not allowed to board’ most concisely indicates the cause for the main event (‘created ruckus’) and should be annotated. Selecting a shorter phrase such as ‘not allowed’ or ‘board’ do not fully capture the cause.
    \item If a document does not contain any information about the cause or reason for the main event, no annotation is required.
\end{itemize}

\section{Corpus Statistics Across Seven Indian Newspapers} \label{appen3}
Table \ref{table10} presents the corpus statistics across seven Indian newspapers, including both national and regional sources. The corpus spans a temporal range from 2001 to 2019 and contains 5.24 million reports. The diversity of sources, along with their varied geographic scope, enhances the overall diversity of the dataset.
\begin{table}[hbt!]
\centering
\resizebox{\columnwidth}{!}{%
\large
\begin{tabular}{l|c|c|c}
\hline
\multicolumn{1}{c|}{\textbf{Newspaper}} & \textbf{Spatial Scope} & \textbf{Temporal Scope} & \textbf{\# Reports} \\ \hline
\textbf{Times of India} & National & 2001 - 2019 & 2,347,370 \\ 
\textbf{The Hindu} & National & 2009 - 2019 & 1,240,781 \\ 
\textbf{The Pioneer} & National & 2011 - 2019 & 328,228 \\ 
\textbf{Economic Times} & National & 2001 - 2019 & 1,187,350 \\ 
\textbf{Assam Tribune} & Regional & 2010 - 2019 & 113,615 \\ 
\textbf{Kashmir Observer} & Regional & 2012 - 2019 & 21,761 \\ 
\textbf{Incredible Orissa} & Regional & 2015 - 2019 & 2,857 \\ \hline
 & \multicolumn{1}{l|}{} & TOTAL & 5,241,962 \\ \hline
\end{tabular}
}%
\caption{National and regional newspapers from India in the corpus.}
\label{table10}
\end{table}

\section{ICR During Coder Training Step}\label{appen4}
Table \ref{table11} presents the ICR measured using Krippendorff’s alpha across three rounds of coder training. The results show a consistent improvement in agreement for all 5Ws. While initial ICR values was relatively low, substantial gains are observed by Round 3, where all Ws achieve strong agreement levels (ICR $\ge $0.8).This demonstrates that the coders developed a reliable understanding of the annotation task before proceeding to annotating EVENT5Ws.
% \usepackage{multirow}
\begin{table}[t]
\centering
\resizebox{\columnwidth}{!}{%
\footnotesize
\begin{tabular}{l|lll}
\hline
\multicolumn{1}{c|}{\multirow{2}{*}{\textbf{5Ws}}} & \multicolumn{3}{c}{\textbf{Krippendorff’s Alpha ($\alpha$)}} \\ \cline{2-4} 
\multicolumn{1}{c|}{} & \multicolumn{1}{c|}{\textbf{Round   1}} & \multicolumn{1}{c|}{\textbf{Round   2}} & \textbf{Round   3} \\ \hline
\textbf{Where} & \multicolumn{1}{c|}{0.45} & \multicolumn{1}{c|}{0.59} & 0.84 \\ \hline
\textbf{When} & \multicolumn{1}{c|}{0.33} & \multicolumn{1}{c|}{0.64} & 0.89 \\ \hline
\textbf{Who} & \multicolumn{1}{c|}{0.78} & \multicolumn{1}{c|}{0.79} & 0.89 \\ \hline
\textbf{What} & \multicolumn{1}{c|}{0.69} & \multicolumn{1}{c|}{0.74} & 0.87 \\ \hline
\textbf{Why} & \multicolumn{1}{c|}{0.32} & \multicolumn{1}{c|}{0.66} & 0.80 \\ \hline
\end{tabular}}
\caption{ICR in different rounds of training.}
\label{table11}
\end{table}

\section{Prompt Examples} \label{appen5}
\begin{figure*}[t]
\centering
\includegraphics[width=\linewidth]{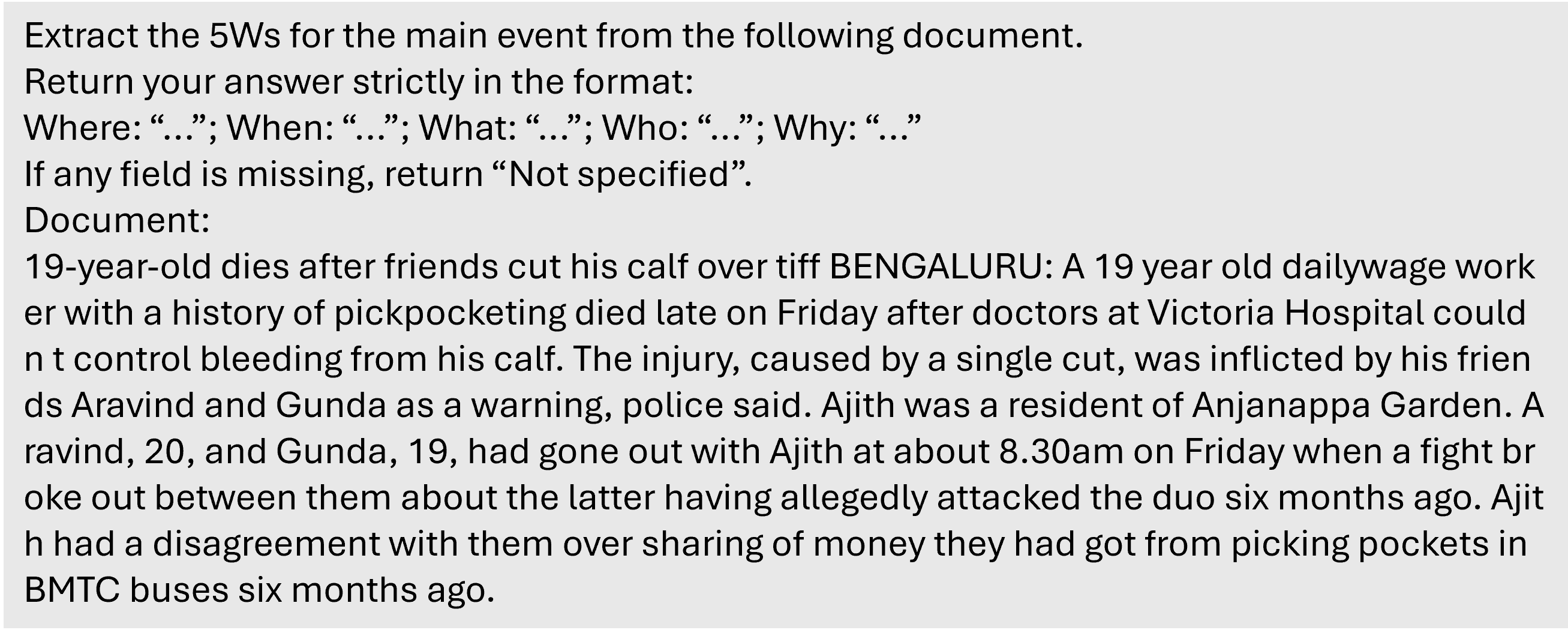}
\caption{Zero-shot prompting example.}
\label{figure7}
\end{figure*}

\begin{figure*}[t]
\centering
\includegraphics[width=\linewidth]{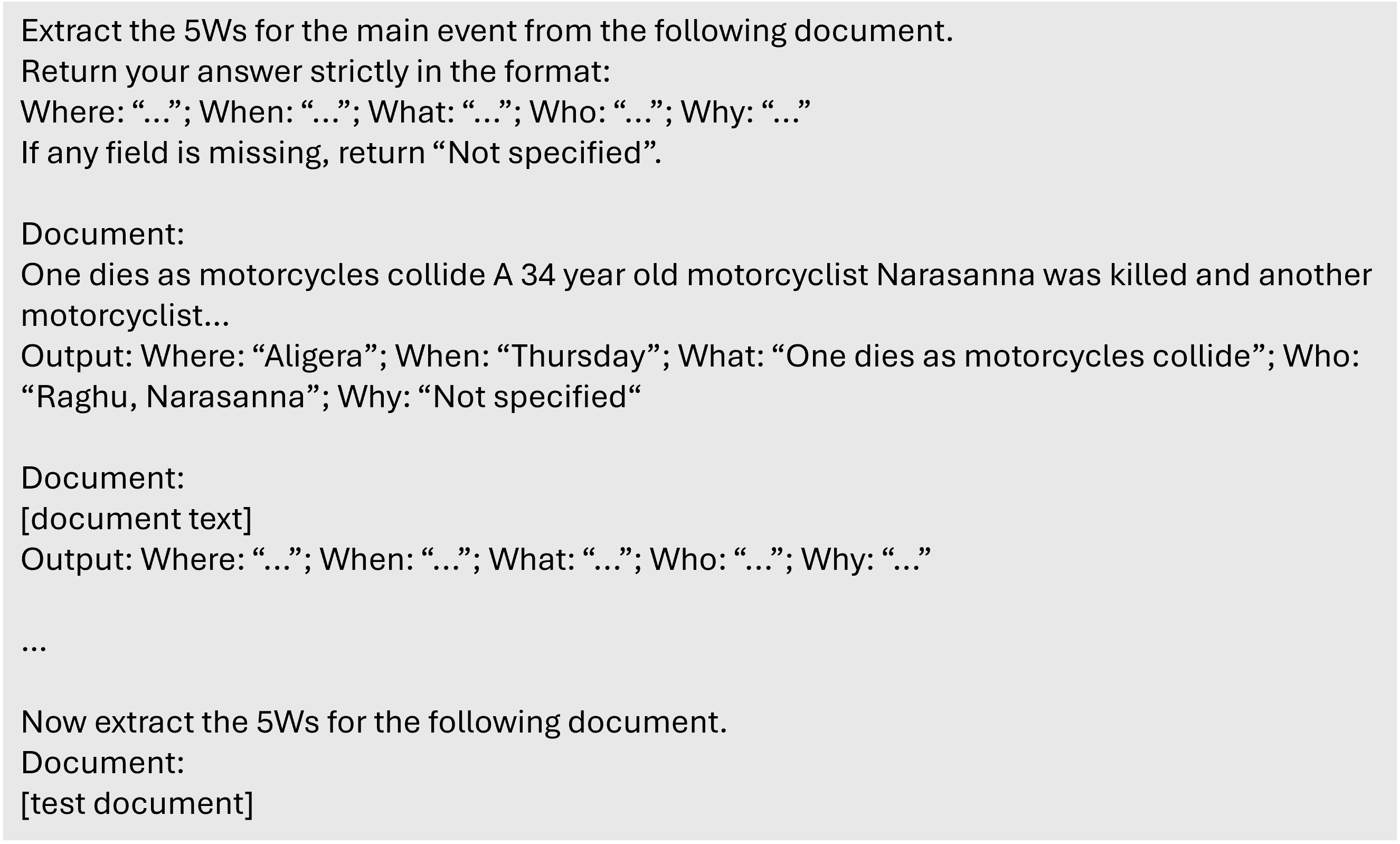}
\caption{Five-shot prompting example.}
\label{figure8}
\end{figure*}

Figure \ref{figure7} and Figure \ref{figure8} present the examples of the prompts used for zero-shot and five-shot settings, respectively. We format all prompts using the chat template specific to each model we evaluate to ensure consistency. In the zero-shot setting, the model is directly instructed to extract the 5Ws from the input document, whereas in the five-shot setting, the prompt includes five examples to guide the model before it is instructed to extract the 5Ws for a test document.

\iftaclpubformat

\onecolumn
% \section{Author/Affiliation Options as set forth by MIT Press}
% \label{sec:authorformatting}

% Option 1. Author’s address is underneath each name, centered.

% \begin{quote}\centering
%   \begin{tabular}{c}
%     \textbf{First Author} \\
%     First Affiliation \\
%     First Address 1 \\
%     First Address 2 \\
%     \texttt{first.email@example.com}
%   \end{tabular}
%   \ 
%   \begin{tabular}{c}
%     \textbf{Second Author} \\
%     Second Affiliation \\
%     Second Address 1 \\
%     Second Address 2 \\
%     \texttt{second.email@example.com}
%   \end{tabular}

%   \begin{tabular}{c}
%     \textbf{Third Author} \\
%     Third Affiliation \\
%     Third Address 1 \\
%     Third Address 2 \\
%     \texttt{third.email@example.com}
%   \end{tabular}
% \end{quote}

% Option 2. Author’s address is linked with superscript characters to its name,
% author names are grouped, centered.

% \begin{quote}\centering
%     \textbf{First Author$^\diamond$} \quad \textbf{Second Author$^\dagger$} \quad
%     \textbf{Third Author$^\ddagger$}
%     \\ \ \\
%     $^\diamond$First Affiliation \\
%     First Address 1 \\
%     First Address 2 \\
%     \texttt{first.email@example.com}
%      \\ \ \\
%      $^\dagger$Second Affiliation \\
%     Second Address 1 \\
%     Second Address 2 \\
%     \texttt{second.email@example.com}
%      \\ \ \\
%     $^\ddagger$Third Affiliation \\
%     Third Address 1 \\
%     Third Address 2 \\
%     \texttt{third.email@example.com}
% \end{quote}
  
%\fi

\end{document}